\documentclass[letterpaper, 10 pt, conference]{ieeeconf}  

\IEEEoverridecommandlockouts                              

\usepackage{graphics} 
\usepackage{epsfig} 
\usepackage{mathptmx} 
\usepackage{times} 
\usepackage{amsmath} 
\usepackage{amssymb}  

\usepackage{cite}
\usepackage{booktabs}
\usepackage{colortbl}
\usepackage{xcolor}
\usepackage{gensymb}
\usepackage{pifont}   
\usepackage{graphicx}
\usepackage{cuted}    
\usepackage{capt-of}  
\usepackage[hidelinks]{hyperref}
\usepackage{makecell}
\usepackage{bbding}   

\title{\LARGE \bf
VCoT-Grasp: Grasp Foundation Models with Visual Chain-of-Thought Reasoning for Language-driven Grasp Generation
}

\author{
Haoran Zhang$^{1*}$, Shuanghao Bai$^{1*\dag}$, Wanqi Zhou$^1$, Yuedi Zhang$^1$, Qi Zhang$^1$, Pengxiang Ding$^{2,3}$, \\
Cheng Chi$^4$, Donglin Wang$^3$, Badong Chen$^{1}$\Envelope \\
$^1$Xi'an Jiaotong University $^2$Zhejiang University $^3$Westlake University $^4$BAAI
\thanks{$^*$Equal contribution. $^\dag$Project leader. \Envelope~Corresponding author.}}

\begin{document}

\maketitle

\thispagestyle{empty}
\pagestyle{empty}

\begin{strip}
  \vspace{-2.5\baselineskip} 
  \centering
  \includegraphics[width=\textwidth]{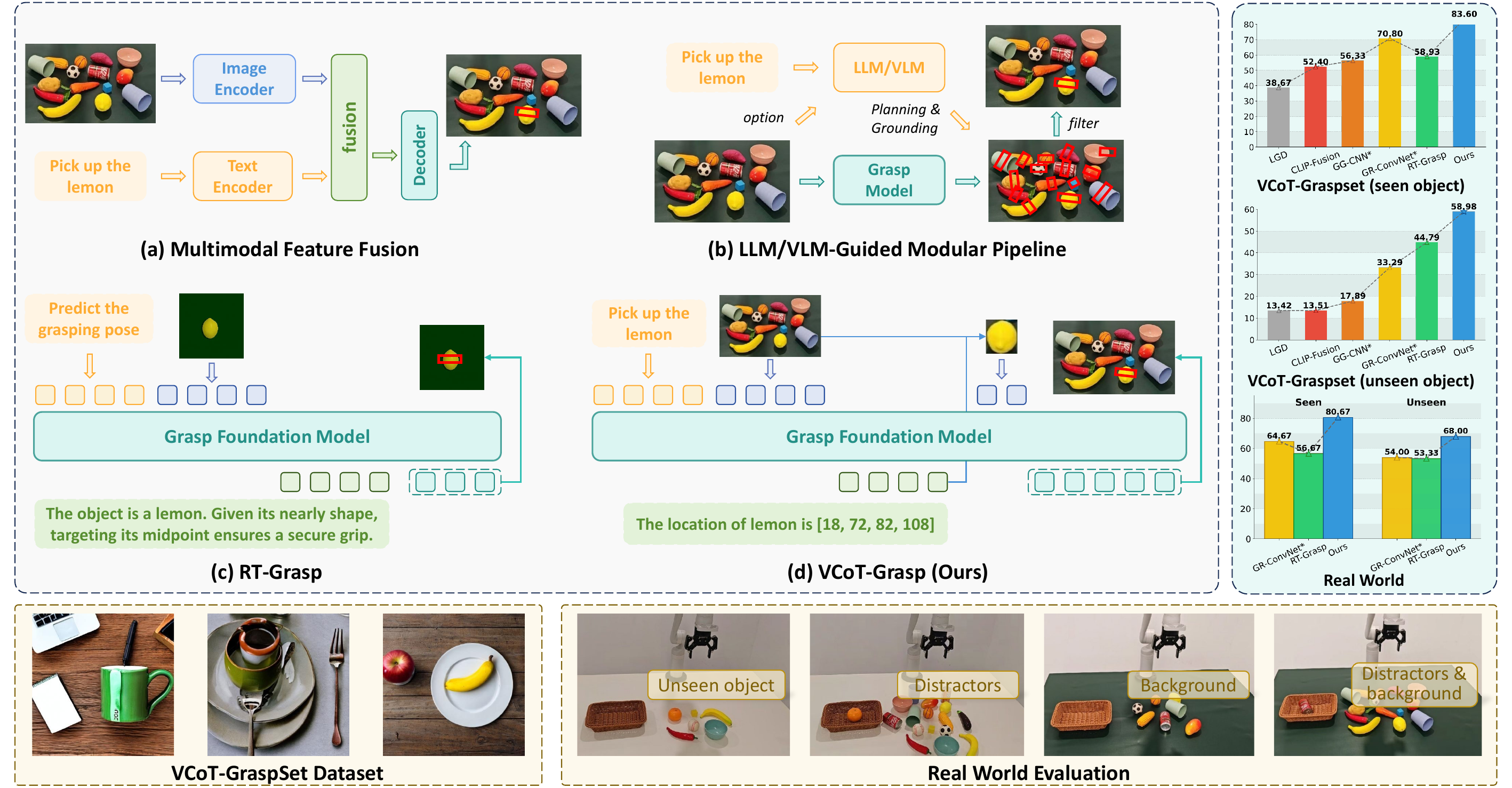}
  \captionof{figure}{Diverging from prior language-driven grasp detection/generation approaches including (a) end-to-end multimodal feature fusion methods~\cite{xu2023joint, nguyen2024language}, (b) LLM/VLM-guided modular pipelines~\cite{tang2023graspgpt}, and (c) end-to-end foundation models with language reasoning~\cite{xu2024rt}, our method (d) advocates visual chain-of-thought reasoning, encouraging the model to ``think with images.” It emphasizes visual grounding by localizing regions that contain critical visual cues and dynamically zooming in to capture context at the appropriate granularity. This mechanism leads to superior generalization to unseen objects, backgrounds, and distractors.
  }
  \label{fig:teaser}
  \vspace{0.5\baselineskip}
\end{strip}

\begin{abstract}

Robotic grasping is one of the most fundamental tasks in robotic manipulation, and grasp detection and generation have long been the subject of extensive research. Recently, language-driven grasp generation has emerged as a promising direction due to its practical interaction capabilities. However, most existing approaches either lack sufficient reasoning and generalization capabilities or depend on complex modular pipelines. Moreover, current grasp foundation models tend to overemphasize dialog and object semantics, resulting in inferior performance and restriction to single-object grasping.
To maintain strong reasoning ability and generalization in cluttered environments, we propose VCoT-Grasp, an end-to-end grasp foundation model that incorporates visual chain-of-thought reasoning to enhance visual understanding for grasp generation. VCoT-Grasp adopts a multi-turn processing paradigm that dynamically focuses on visual inputs while providing interpretable reasoning traces.
For training, we refine and introduce a large-scale dataset, VCoT-GraspSet, comprising 167K synthetic images with over 1.36M grasps, as well as 400+ real-world images with more than 1.2K grasps, annotated with intermediate bounding boxes. Extensive experiments on both VCoT-GraspSet and real robot demonstrate that our method significantly improves grasp success rates and generalizes effectively to unseen objects, backgrounds, and distractors. Our code and dataset can be found at \url{https://zhanghr2001.github.io/VCoT-Grasp.github.io}.

\end{abstract}

\section{Introduction}

Grasping is a fundamental capability in tabletop manipulation and serves as the basis for downstream tasks such as placing, rearranging, or relocating objects. Traditional grasp detection methods~\cite{redmon2015real, kumra2017robotic, morrison2018closing, kumra2020antipodal, fang2020graspnet, ainetter2021end, fang2023anygrasp} typically rely on Convolutional Neural Networks (CNNs) to extract visual and spatial features, or on generative models such as Variational Auto-Encoders (VAEs)~\cite{mousavian20196, sundermeyer2021contact}, which propose feasible grasps across the entire scene. More recently, transformer-based approaches~\cite{xiong2024hmt, fan2025miscgrasp} have been introduced. While these methods can generate high-quality grasps, they lack the ability to focus on a specific target object, often requiring additional post-processing to filter relevant grasps. This limitation reduces semantic guidance and poses challenges for seamless human–robot interaction in real-world environments.

To overcome these limitations, language-driven grasping has emerged, enabling object-specific and instruction-guided manipulation. Some approaches incorporate language as an additional modality for multimodal feature fusion~\cite{xu2023joint, nguyen2024language, yang2024attribute, vuong2024language, jin2025reasoning}, while others exploit large language and vision-language models for planning and grounding to filter or rank grasp candidates~\cite{lu2023vl, tang2023graspgpt, qian2025thinkgrasp, tziafas2025towards, deshpande2025graspmolmo, du2025finegrasp}. By integrating natural language understanding, these methods enhance practicality in real-world applications.
However, the former lacks explicit reasoning capabilities, while the latter often depends on modular, cascading pipelines, resulting in larger and more complex systems with higher computational and memory costs. Both approaches still exhibit limited generalization. Given that foundation models have already demonstrated strong textual and visual generalization, we aim to develop an end-to-end reasoning foundation model for grasp detection and generation.

Recently, RT-Grasp~\cite{xu2024rt} introduced the first end-to-end foundation model specifically designed for language-driven grasp generation. It demonstrates impressive language reasoning ability to infer and refine predicted grasps. However, RT-Grasp still suffers from several limitations, including inefficient training, applicability restricted to single-object scenarios, and inferior performance that is even slightly worse than traditional methods. We attribute these shortcomings to its overemphasis on dialog and object semantics rather than direct visual understanding, which makes it particularly ineffective in complex and cluttered environments.

To this end, inspired by recent ``think with images” techniques~\cite{shao2024visual, zheng2024instruction, wang2025vgr}, we propose VCoT-Grasp, an end-to-end VLM-based framework that incorporates visual chain-of-thought reasoning to enhance visual understanding for grasp generation. The key insight is that humans interpret complex visual information by selectively focusing on salient regions or fine details, rather than processing the entire scene uniformly. In contrast, most robotic grasp foundation models handle aligned image contexts at a fixed granularity, thereby constraining efficiency and flexibility. To emulate human-like efficient reasoning, models must identify critical regions containing essential visual cues and dynamically zoom in to capture context at the appropriate scale. 
Specifically, given an image and a language instruction, the model first identifies the target object and predicts its bounding box. The corresponding region is then cropped and resized to a normalized scale, and fed back into the model to generate the final grasp prediction. This multi-turn in-context paradigm improves visual understanding and enables finer-grained reasoning. Moreover, the potential of multi-turn in-context learning and the advantages of chain-of-thought reasoning remain largely unexplored in robotic foundation models, and our work bridges this gap.

Foundation models typically require large-scale datasets for effective training. However, existing large-scale grasp datasets are predominantly synthetic and automatically annotated, leading to substantial noise and limited accuracy. To address this issue, we refine and recollect a new dataset VCoT-GraspSet based on~\cite{vuong2024grasp}. In total, the dataset comprises 167K synthetic images with corresponding grasp annotations and more than 400 real-world images with over 1.2K annotated grasps.
Our main contributions are as follows:
\begin{itemize}
\item We propose VCoT-Grasp, an end-to-end foundation model that combines language-driven grasp generation with visual chain-of-thought reasoning, improving visual understanding, grasp quality, and generalization.
\item We present VCoT-GraspSet, a refined grasping dataset comprising 167K synthetic images with over 1.36M grasps and 400+ real-world images with more than 1.2K grasps. Each sample includes an image, grasp annotations, and intermediate bounding boxes that serve as chain-of-thought context.
\item Extensive experiments on both VCoT-GraspSet and real-world scenarios demonstrate that VCoT-Grasp produces high-quality grasps and generalizes effectively to unseen objects, backgrounds, and distractors.
\end{itemize}

\section{Related Work}

\noindent \textbf{Robotic Grasp.}
Grasp detection~\cite{jiang2011efficient} and grasp generation~\cite{mousavian20196} are two closely related tasks in robotic grasping, differing mainly in whether generative models are used to produce grasp candidates. The objective of grasp detection/generation is to generate grasp poses that can be directly executed by robotic manipulators to stably grasp objects.
Learning-based methods typically follow either a sample-and-evaluate strategy or a regression-based formulation, often trained under a behavior cloning objective~\cite{newbury2023deep}. Traditional approaches rely heavily on CNNs~\cite{redmon2015real, kumra2017robotic, morrison2018closing, kumra2020antipodal, fang2020graspnet, ainetter2021end, fang2023anygrasp}. For instance, GG-CNN~\cite{morrison2018closing} and GR-ConvNet~\cite{kumra2020antipodal} treat each image pixel as a potential grasp center and predict grasp width, angle, and success score for every pixel. GraspNet-Baseline~\cite{fang2020graspnet} and FGC-GraspNet~\cite{lu2022hybrid} extend this idea to point clouds by considering each point as a candidate contact, sampling approach vectors in Euclidean space, and regressing the remaining grasp parameters.
Generative approaches~\cite{mousavian20196, sundermeyer2021contact, huang2025robograsp} such as 6-DoF GraspNet~\cite{mousavian20196} leverage a variational autoencoder to sample candidate grasps, followed by iterative evaluation and refinement. More recently, transformer-based architectures~\cite{xiong2024hmt, fan2025miscgrasp} have been proposed. 
While these methods achieve strong in-distribution performance, they struggle to generalize to unseen and out-of-distribution (OOD) scenarios and lack the ability to select target objects through practical semantic reasoning, which limits their applicability in real-world human–robot interaction. To address these limitations, we propose VCoT-Grasp, which bridges this gap through visual chain-of-thought reasoning.

\noindent \textbf{Language-driven Grasping.}
Driven by the practical advantages of semantics and the success of large language models, language-driven grasping has gained increasing attention. A common strategy in this line of research is multimodal representation learning with feature fusion~\cite{xu2023joint, nguyen2024language, yang2024attribute, vuong2024language, jin2025reasoning, noh2025graspsam}. For example, LGrasp6D~\cite{nguyen2024language} and Generic~\cite{yang2024attribute} encode image and text modalities separately, fuse them via cross-attention, and then use a decoder (e.g., diffusion models) to generate grasps. However, these methods primarily focus on data fitting rather than reasoning and remain ineffective in OOD scenarios.
Another line of work leverages pretrained grasp models to first generate grasp candidates, followed by LLMs or VLMs for reasoning or visual grounding to rank or filter grasps, ultimately selecting the most confident one~\cite{lu2023vl, tang2023graspgpt, qian2025thinkgrasp, tziafas2025towards, deshpande2025graspmolmo, du2025finegrasp}. Such approaches form modular pipelines, leading to larger and more complex robotic systems with increased computational and memory demands, while the pretrained grasp model inherently constrains the upper bound of performance.
More recently, several methods~\cite{xu2024rt, chu2025graspcot} have explored end-to-end fine-tuning with foundation models for grasp generation. While these approaches exhibit strong text-based reasoning and grasp-correction abilities, systems such as RT-Grasp~\cite{xu2024rt} tend to overemphasize dialog and object semantics. This often leads to suboptimal grasping performance and reduced effectiveness in cluttered or visually ambiguous environments.
Different from these, we propose a simple yet effective framework, VCoT-Grasp, which introduces visual chain-of-thought reasoning to enhance visual understanding and substantially improve OOD generalization in complex environments.

\section{Method}

\subsection{Problem Formulation}

In this work, we define the robotic grasping problem as the task of identifying an antipodal grasp, oriented perpendicular to a planar surface, given an RGB image of the scene together with a corresponding natural language instruction that specifies the target object. We adopt the widely used rectangle grasp representation~\cite{jiang2011efficient, kumra2020antipodal}, in which each grasp configuration is parameterized by five values:
\begin{equation}
    g = [x, y, w, h, \theta]
\end{equation}
where $[x, y]$ denote the pixel coordinates of the grasp center within the image plane, while $[w, h]$ correspond to the width and height of the grasp rectangle, representing the physical gripper width and finger width, respectively. The final parameter, $\theta \in [0^\circ, 180^\circ]$, specifies the rotation angle of the gripper relative to the horizontal axis, thereby determining the grasp orientation.
A language-driven grasp generation model $\pi$ maps an image observation $O \in \mathbb{R}^{H \times W \times 3}$ and a language instruction $l$ to a grasp $g$, formulated as:
\begin{equation}
g = \pi(O, l).
\label{grasp_formulate}
\end{equation}

Our VCoT-Grasp model introduces visual chain-of-thought reasoning, enabling two key capabilities: visual understanding and grasp generation.
Firstly, we equip the model with the ability to localize the target object for grasping by predicting its bounding box image as an intermediate reasoning step:
\begin{align}\label{step1}
b &= \pi(O, l_d),
\end{align}
where $l_d$ denotes the detection instruction of the target object. In this step, the model distinguishes the target from irrelevant objects and provides a coarse-grained location.

Next, the bounding box is used to crop and resize a square region of the image, yielding the bounding box image $O_b$ that effectively zooms in on the region of interest. The model takes as input both the original and localized images to generate a refined grasp pose:
\begin{align}\label{step2}
g &= \pi(O, O_b, l_g),
\end{align}
where $l_g$ denotes the grasp instruction for the target object. With zoomed-in image, the model can focus on fine-grained position adjustment, enabling high-quality grasp prediction.

\subsection{Dataset}

\begin{figure}[tbp]
\centering
\includegraphics[width=\columnwidth]{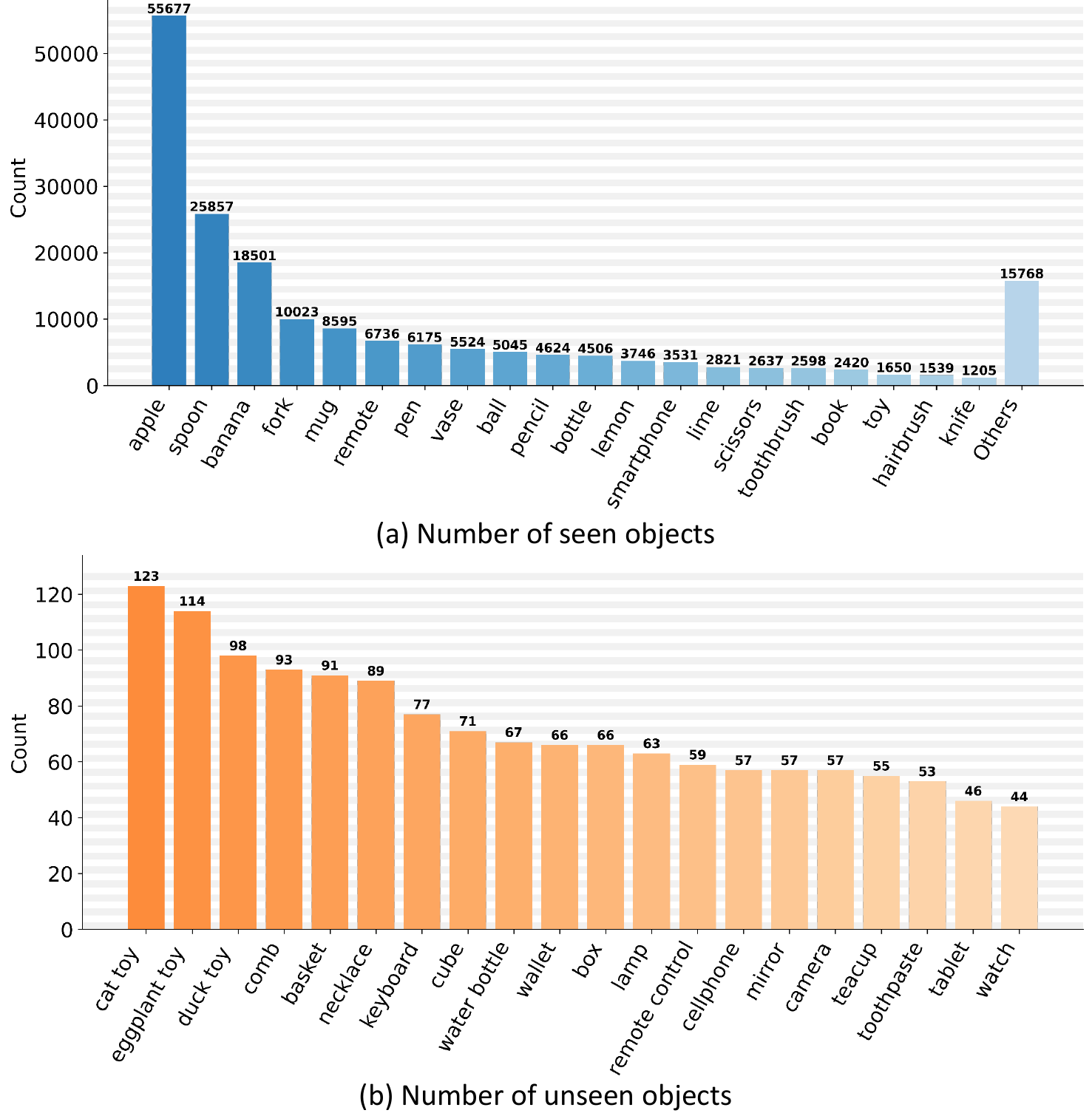}
\caption{Dataset Statistics. We report the number of (a) seen and (b) unseen objects. In the seen scenario, Others denotes the aggregated count of the remaining 347 categories.
}
\label{fig:dataset}
\end{figure}

So far, most existing large-scale grasp datasets are generated automatically. For example, Grasp Anything~\cite{vuong2024grasp} contains approximately 1M synthetic images and 33M grasp poses across diverse scenes and object categories. However, its text-to-image generation and segmentation pipeline can produce distorted objects, inaccurate masks, and mismatched object names, resulting in unreliable bounding boxes and grasp annotations.
Training directly on such noisy datasets severely compromises real-world deployment, highlighting the need for refinement and recollection. Based on~\cite{vuong2024grasp}, we leverage the open-vocabulary detection model YOLO-World~\cite{cheng2024yolo} to reprocess the data. Specifically, we compare the bounding boxes produced by YOLO-World with those from the original dataset and filter out objects with an intersection-over-union (IoU) lower than 0.25. This step effectively removes low-quality or distorted objects that fail to be detected, as well as mismatched objects and names thus unreliable grasp poses.
This step removes 86\% of original objects and keeps the most high quality part of the dataset.
The filtered data are further validated through crowdsourcing to ensure that no distorted objects remain and that bounding boxes and grasp poses are correctly annotated.

The resulting VCoT-GraspSet contains over 190K images (we denote the number of images as the number of samples) and over 1.36M grasps spanning 388 object categories. Among these, 367 categories are designated as seen objects, split into a training set with over 186K samples and a test set with about 3K samples, while the remaining 21 categories are reserved as unseen objects with approximately 1.5K samples for testing. The dataset statistics are summarized in Figure~\ref{fig:dataset}.
We additionally collect a real-world dataset containing 450 images of 30 objects and over 1.2K manually annotated grasps and bounding boxes. These objects are evenly divided into 15 seen and 15 unseen categories.

\subsection{VCoT-Grasp Architecture}
The overall framework of VCoT-Grasp is illustrated in Figure~\ref{fig:model}. VCoT-Grasp is an end-to-end model that generates grasp poses through an intermediate visual reasoning step. It is built upon the pretrained VLM PaliGemma2~\cite{steiner2024paligemma} and consists of four main components:

\begin{figure}[tbp]
\centering
\includegraphics[width=\columnwidth]{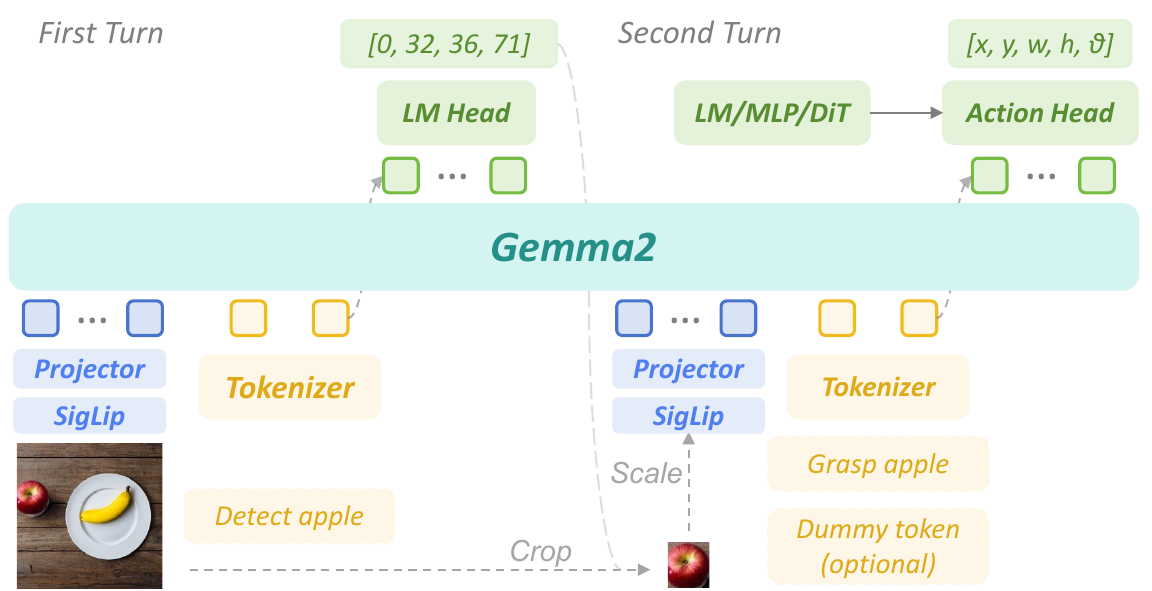}
\caption{Overall framework of VCoT-Grasp. Our grasp model architecture is built on the Paligemma-3B VLM~\cite{steiner2024paligemma}, which takes as input projected visual embeddings and tokenized task instructions, and employs multi-turn learning to predict both the location tokens of the target object and the grasp pose tokens. We evaluate multiple action head designs, where these decoders exploit fine-grained visual information to generate grasp poses either in an autoregressive or regression manner.}
\label{fig:model}
\end{figure}

\noindent \textbf{Image Encoder.} The observation $O$, which can be either the gripper camera view or the static camera view, is encoded into image embeddings using a frozen SigLIP image encoder~\cite{zhai2023sigmoid}.

\noindent \textbf{Visual Projector.} A linear projection module that aligns SigLIP’s visual features with the embedding space of the language model.

\noindent \textbf{LLM Backbone.} We adopt Gemma2~\cite{team2024gemma} as the backbone for both language understanding and grasp decoding, together with its native tokenizer for text processing.
To represent the five-parameter grasp box, we explore two forms: (1) continuous values, where the LLM's output features are projected to regress grasp labels, and (2) discrete tokens, where the parameters are expressed as tokens from the language model’s vocabulary and predicted via the next-token-prediction paradigm.

\noindent \textbf{Action Head.} The action head serves as a decoder that generates grasp poses from the final-layer hidden representations. In this work, we systematically investigate different design choices for the action head under both formulations, considering four variants: (1) a multilayer perceptron (MLP) head, (2) a diffusion head based on the Diffusion Transformer (DiT)~\cite{peebles2023scalable, chi2023diffusion}, and (3–4) direct reuse of the language model’s output head (LM head) with either pretrained tokens or newly introduced tokens.
When employing the MLP or diffusion head, we insert several dummy tokens into the input prompts and use their final-layer features to predict grasp. For the LM head, each dimension of the grasp is discretized into 1024 tokens in language model's vocabulary. A common practice is either to introduce new tokens or to override the least frequently used tokens. In this work, we explore reusing PaliGemma’s pretrained position tokens, and additionally evaluate the use of our newly added tokens.


\subsection{Training Strategy}

In our dataset, each image $O$ may contain multiple objects along with their grasp-related textual descriptions. Consequently, a single image can correspond to multiple image–bounding box pairs $b$ and grasp poses $g$. 
Depending on the design of the action head, action prediction can be formulated either as continuous regression using a MLP or diffusion head, or as discrete classification using a LM head. The prediction loss can be expressed as:
\begin{align}\label{loss}
\mathcal{L}_{\text {total}}=\mathcal{L}_g(g, \hat{g})+\lambda \mathcal{L}_b(b, \hat{b}),
\end{align}
where $\hat{g}$ and $\hat{b}$ denote the predicted grasp pose and bounding box labels, respectively. The grasp loss $\mathcal{L}_g$ corresponds to the L1 loss for the MLP head, the mean-squared error loss for the diffusion head, and the cross-entropy loss for the LM head. The box loss $\mathcal{L}_b$ is a cross-entropy loss. The weighting factor $\lambda$ is fixed to 1.0.

\section{Experiments}

In this section, we aim to answer three key questions about VCoT-Grasp. (1) How well does the model perform on standard grasping benchmarks and real-robot manipulation tasks? (Section~\ref{subsec:eval_sim} and \ref{subsec:eval_real}) (2) Which components, including visual chain-of-thought reasoning, the choice of action head, the training schedule, and data scale, are essential to its overall performance? (Section~\ref{subsec:ablation}) (3) Does the model exhibit zero-shot generalization, transferring from VCoT-GraspSet to real robots while remaining robust to background changes and distractor objects? (Section~\ref{subsec:generalization}).

\subsection{Evaluation on VCoT-GraspSet}
\label{subsec:eval_sim}

\begin{table*}[t]
\centering
\caption{Real-world grasping success rates (successes/trials) on 15 seen objects.}
\label{tab:real_world_seen}
\vskip -0.05in
\resizebox{\textwidth}{!}{
\begin{tabular}{lcccccccccccccccc}
\toprule
Method & \rotatebox{90}{Coke} & \rotatebox{90}{Fanta} & \rotatebox{90}{Orange} & \rotatebox{90}{Lemon} & \rotatebox{90}{Carrot} & \rotatebox{90}{Red Pepper} & \rotatebox{90}{Eggplant} & \rotatebox{90}{Baseball} & \rotatebox{90}{Soccer Ball} & \rotatebox{90}{Blue Cube} & \rotatebox{90}{Red Cube} & \rotatebox{90}{Red Bowl} & \rotatebox{90}{Green Bowl} & \rotatebox{90}{White Cup} & \rotatebox{90}{Blue Cup} & Avg. \\
\midrule
GR-ConvNet~\cite{kumra2020antipodal} + CLIP~\cite{radford2021learning} & 6/10 & 9/10 & 10/10 & 10/10 & 9/10 & 10/10 & 4/10 & 9/10 & 8/10 & 9/10 & 10/10 & 0/10 & 3/10 & 0/10 & 0/10 & 0.65 \\
RT-Grasp~\cite{xu2024rt}           & 2/10 & 6/10 & 7/10 & 8/10 & 7/10 & 5/10 & 5/10 & 10/10 & 4/10 & 10/10 & 6/10 & 7/10 & 7/10 & 1/10 & 0/10 & 0.57 \\
\rowcolor{gray!9.0} VCoT-Grasp (Ours)               & 9/10 & 9/10 & 6/10 & 8/10 & 9/10 & 10/10 & 10/10 & 10/10 & 9/10 & 10/10 & 8/10 & 7/10 & 3/10 & 6/10 & 7/10 & \textbf{0.81} \\
\bottomrule
\end{tabular}
}
\end{table*}

\begin{table*}[ht]
\centering
\caption{Real-world grasping success rates (successes/trials) on another set of 15 household unseen objects.}
\label{tab:real_world_unseen}
\vskip -0.05in
\resizebox{\textwidth}{!}{
\begin{tabular}{lcccccccccccccccc}
\toprule
Method & \rotatebox{90}{Spirit} & \rotatebox{90}{Banana} & \rotatebox{90}{Mango} & \rotatebox{90}{Corn} & \rotatebox{90}{Potato} & \rotatebox{90}{Red Potato} & \rotatebox{90}{Green Pepper} & \rotatebox{90}{Basketball} & \rotatebox{90}{Tennis Ball} & \rotatebox{90}{Green Cube} & \rotatebox{90}{Yellow Cube} & \rotatebox{90}{White Bowl} & \rotatebox{90}{Blue Bowl} & \rotatebox{90}{Green Cup} & \rotatebox{90}{Red Cup} & Avg. \\
\midrule
GR-ConvNet~\cite{kumra2020antipodal} + CLIP~\cite{radford2021learning} & 6/10 & 7/10 & 5/10 & 9/10 & 10/10 & 8/10 & 3/10 & 10/10 & 6/10 & 4/10 & 7/10 & 2/10 & 3/10 & 1/10 & 0/10 & 0.54 \\
RT-Grasp~\cite{xu2024rt}           & 3/10 & 5/10 & 2/10 & 9/10 & 5/10 & 5/10 & 5/10 & 10/10 & 9/10 & 7/10 & 4/10 & 6/10 & 7/10 & 3/10 & 0/10 & 0.53 \\
\rowcolor{gray!9.0} VCoT-Grasp (Ours)               & 8/10 & 7/10 & 9/10 & 10/10 & 6/10 & 3/10 & 10/10 & 10/10 & 10/10 & 9/10 & 9/10 & 6/10 & 3/10 & 1/10 & 1/10 & \textbf{0.68} \\
\bottomrule
\end{tabular}
}
\end{table*}

\begin{table}[htbp]
\caption{Comparison with language-conditioned grasp models on VCoT-GraspSet. Avg. denotes the harmonic mean. Best and second-best are shown in bold and underline, respectively.}
\centering
\vskip -0.05in
\begin{tabular}{
>{\raggedright\arraybackslash}m{3.5cm}|
>{\centering\arraybackslash}m{1cm}
>{\centering\arraybackslash}m{1cm}
>{\centering\arraybackslash}m{1cm}
}
\toprule
Method & Seen & Unseen & Avg. \\
\midrule
LGD~\cite{vuong2024language}  & 38.67 & 13.42 & 19.93 \\
CLIP-Fusion~\cite{xu2023joint} & 52.40 & 13.51 & 21.48 \\
GG-CNN~\cite{morrison2018closing} + CLIP~\cite{radford2021learning} & 56.33 & 17.89 & 27.16 \\
GR-ConvNet~\cite{kumra2020antipodal} + CLIP~\cite{radford2021learning} & 70.80 & 33.29 & 45.29 \\
RT-Grasp~\cite{xu2024rt} & 58.93 & 44.79 & 50.90 \\
\rowcolor{gray!9.0} VCoT-Grasp (Ours) &  &  &  \\
\rowcolor{gray!9.0} $~~~~$ w/ MLP  Head & \underline{73.37} & \underline{52.25} & \underline{61.03} \\
\rowcolor{gray!9.0} $~~~~$ w/ LM  Head & \textbf{83.60} & \textbf{58.98} & \textbf{69.16} \\
\bottomrule
\end{tabular}
\label{table:main_result}
\end{table}

\noindent \textbf{Baselines.}
We compare our proposed VCoT-Grasp with a diverse set of grasp detection and generation baselines, which can be grouped into three categories:
(1) multimodal feature fusion methods, including GG-CNN~\cite{morrison2018closing}, GR-ConvNet~\cite{kumra2020antipodal}, and LGD~\cite{vuong2024language};
(2) LLM/VLM-guided modular pipelines, represented by CLIP-Fusion~\cite{xu2023joint}; and
(3) end-to-end foundation models, represented by RT-Grasp~\cite{xu2024rt}.
Among these, GG-CNN and GR-ConvNet are originally designed without language input. We extend their architectures by integrating CLIP’s text encoder~\cite{radford2021learning} and fusing the resulting language embeddings with their visual features, thereby enabling language-driven grasp prediction.
For RT-Grasp, we replace its original LLaVA backbone~\cite{liu2023visual} with the Paligemma backbone~\cite{steiner2024paligemma} and train it under the non-reasoning-B setting described in the original paper~\cite{xu2024rt}.
For fairness, all methods are trained on the VCoT-GraspSet training set using their default hyperparameters as specified in the original papers.

\noindent \textbf{Metrics.}
We evaluate all methods on the test fold of the VCoT-GraspSet dataset following the standard evaluation protocol. A predicted grasp is considered successful if it attains an intersection-over-union (IoU) greater than 0.25 with at least one ground-truth grasp and its orientation deviation is within $30^\circ$.

\noindent \textbf{Implementation Details.}
We adopt the 3B variant of PaliGemma2~\cite{steiner2024paligemma} with an input resolution of $224\times224$ as the base VLM. During training, the image encoder is kept frozen to preserve pretrained visual representations, while all remaining components are updated. The model is trained on over 186K samples from the training split of VCoT-GraspSet for three epochs with an effective batch size of 32 on four NVIDIA A100 GPUs. We use the AdamW optimizer~\cite{kingma2014adam} with a learning rate of $2\times10^{-5}$, along with a cosine learning rate scheduler and a warm-up ratio of 3\%.

\noindent \textbf{Results.} 
As shown in Table~\ref{table:main_result}, we evaluate grasp accuracy on the test split of VCoT-GraspSet. VCoT-Grasp achieves the best performance on both in-distribution seen objects and out-of-distribution unseen objects. The MLP and LM variants outperform prior state of the art by 2.57\% and 12.80\% on seen objects, and by 7.46\% and 14.19\% on unseen objects, respectively. Traditional feature-fusion and VLM-guided modular baselines perform reasonably on seen objects but deteriorate significantly on unseen ones. RT-Grasp benefits from the generalization ability of its foundation model and therefore performs better than these baselines under OOD conditions; however, it still over-relies on linguistic cues and lacks strong visual grounding, resulting in performance that remains inferior to our VCoT-Grasp.

\begin{figure*}[htbp]
\centering
\includegraphics[width=\textwidth]{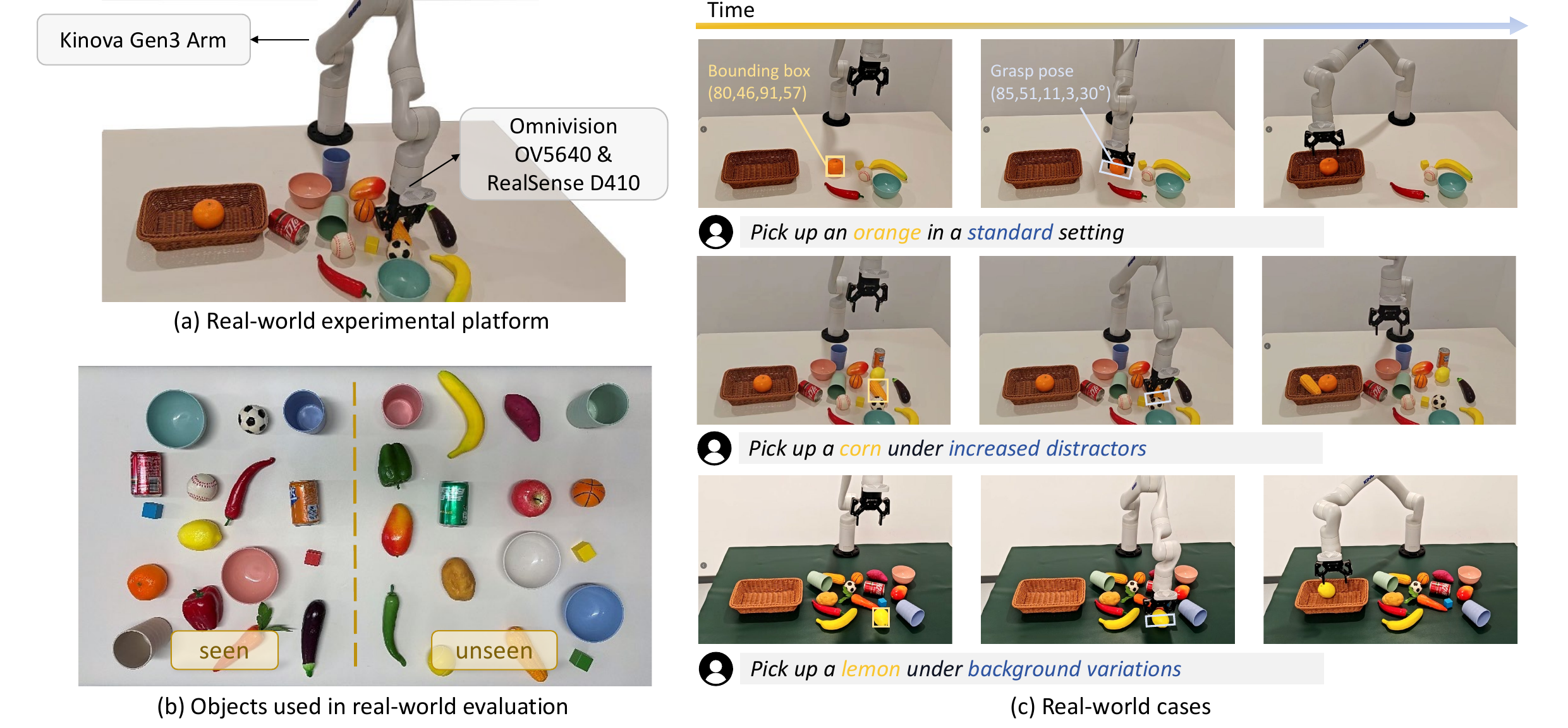}
\caption{Real-world experimental setup. (a) The hardware platform. (b) The 30 household objects used for evaluation, equally divided into seen and unseen categories. (c) Out-of-distribution scenarios with altered backgrounds and additional distractors.}
\label{fig:exp_setup}
\end{figure*}

\subsection{Evaluation in Real-World}
\label{subsec:eval_real}

\noindent \textbf{Setup.}
Our experiments were conducted on a Kinova Gen3 robotic arm equipped with a Robotiq 2F-85 parallel gripper and an Omnivision OV5640 RGB camera, as shown in Figure~\ref{fig:exp_setup} (a). The predicted grasp rectangle was transformed into a 3D pose using camera intrinsics and depth information, and then passed to a motion planner for trajectory generation. To evaluate generalization, we selected 30 everyday household objects spanning a wide range of shapes, sizes, and surface materials. Following the previous setting, the objects were divided into two groups of 15 seen and 15 unseen items, as illustrated in Figure~\ref{fig:exp_setup} (b). 

\noindent \textbf{Baselines.}
Based on the evaluation results on VCoT-GraspSet, we select GR-ConvNet~\cite{kumra2020antipodal} and RT-Grasp~\cite{xu2024rt} as the most competitive baselines for our real-world experiments.

\noindent \textbf{Metrics.}
Each object was tested in 10 different positions and orientations, and performance was measured by grasp success rate. A trial was considered successful if the robotic arm could stably grasp and lift the target object from the table without noticeable collision.

\noindent \textbf{Implementation Details.}
Prior to testing, we constructed a real-world grasp dataset using the seen objects, consisting of approximately 450 RGB images with over 1.2K grasp annotations and bounding boxes per item. All models were fine-tuned on the training split of this dataset to better adapt to real-world objects. 

\noindent \textbf{Results.}
As shown in Table~\ref{tab:real_world_seen} and Table~\ref{tab:real_world_unseen}, we report the grasp success rates of VCoT-Grasp compared with baselines on both seen and unseen objects. Our method consistently achieves the highest success rate across both splits, demonstrating superior robustness and generalization. We also observe that RT-Grasp performs worse than GR-ConvNet, which is consistent with the findings of the original paper~\cite{xu2024rt}, where this limitation is attributed to difficulties in reasoning about unseen object attributes. However, we argue that the weaker performance of RT-Grasp arises from its heavy reliance on dialogue and text-based reasoning, which comes at the expense of visual reasoning ability, leading to inferior results on both seen and unseen objects.


\subsection{Ablation Study}
\label{subsec:ablation}

\noindent \textbf{Effect of Visual Chain-of-Thought Reasoning.}
If the visual chain-of-thought reasoning is removed, the framework degenerates into directly predicting the grasp pose in a single-turn paradigm. As shown in the first four rows of Table~\ref{table:ablation}, eliminating visual chain-of-thought reasoning leads to a substantial performance drop on both seen and unseen test splits. This underscores the importance of intermediate visual reasoning steps, which provide structured guidance and markedly improve grasp prediction accuracy.

\begin{table}[tbp]
\caption{Effect of visual chain-of-thought (VCoT) reasoning and different action heads.}
\label{table:ablation}
\centering
\vskip -0.05in
\begin{tabular}{ccccccc}
\toprule
w/ VCoT & Action Head & Seen & Unseen & Avg. \\
\midrule
\rowcolor{gray!9.0} \checkmark & MLP & 73.37 & 52.25 & 61.03 \\
\ding{55} & MLP & 67.60 & 49.36 & 57.06 \\
\rowcolor{gray!9.0} \checkmark & LM pretrained token & 83.60 & 58.98 & 69.16 \\
\ding{55} & LM pretrained token & 77.29 & 54.07 & 63.63 \\
\checkmark & LM new token & 82.89 & 58.80 & 68.80 \\
\checkmark & Diffusion & 57.50 & 41.29 & 48.07 \\
\bottomrule
\end{tabular}
\end{table}

\noindent \textbf{Effect of Action Heads.}
The choice of action head is a critical component of model design. Table~\ref{table:ablation} reports performance across different action head variants. LM new token and LM pretrained token denote using newly added tokens and reusing PaliGemma’s pretrained location tokens, respectively. Overall, discrete action representations outperform both the MLP and diffusion heads, making them more effective for grasp prediction. Notably, pretrained location tokens, originally designed for object detection in VLM, transfer effectively to grasping task and achieve higher performance than newly introduced tokens. This suggests that positional priors embedded in pretrained VLMs are not only reusable but also beneficial for downstream tasks that require fine-grained spatial reasoning.

\begin{figure}[tbp]
\centering
\includegraphics[width=\columnwidth]{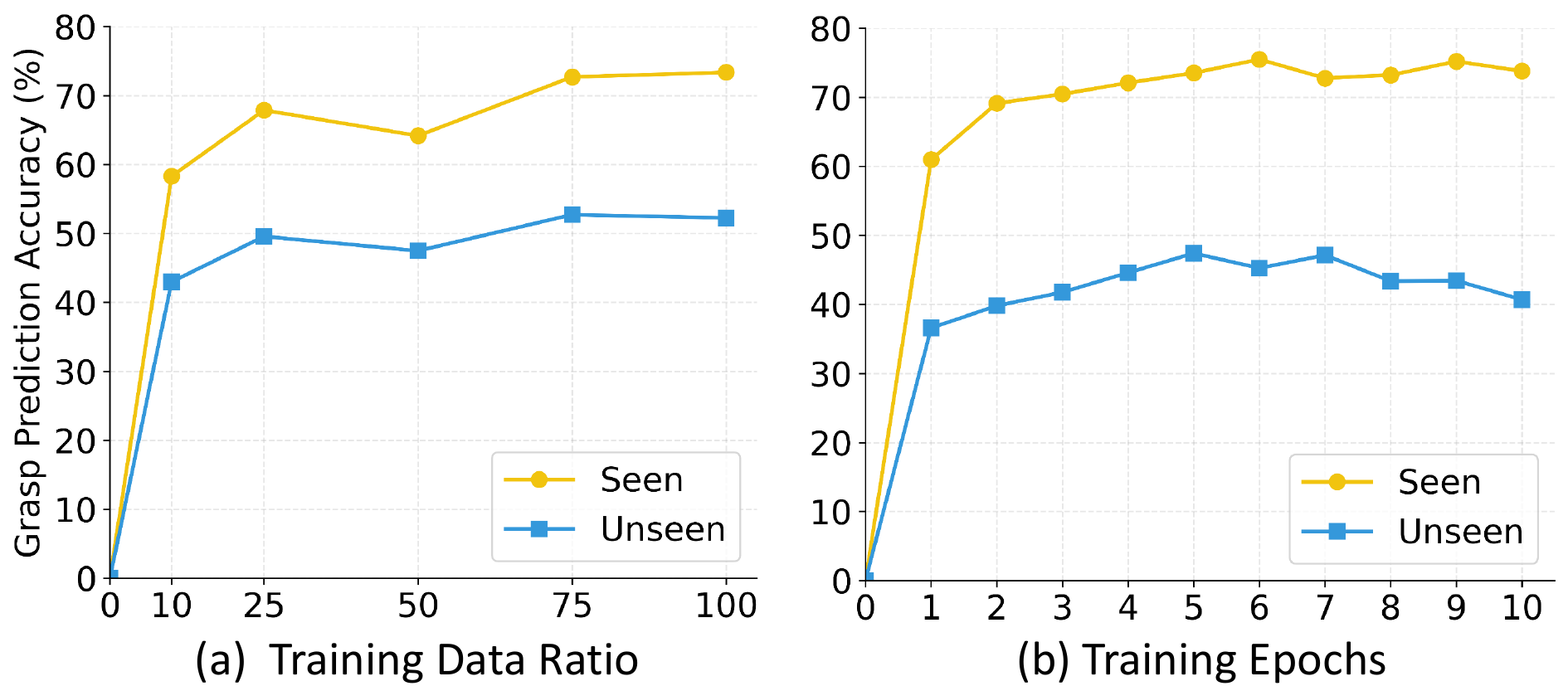}
\caption{(a) Grasp prediction performance as a function of the number of training samples. (b) Effect of training epochs, where overfitting emerges after the fifth epoch.}
\label{fig:data_and_epoch}
\end{figure}

\noindent \textbf{Data Scaling Laws.}
Data diversity is a critical factor in training robust grasping models. To assess its impact, we train multiple models with different proportions of the dataset. As shown in Figure~\ref{fig:data_and_epoch} (a), performance improves consistently as the amount of training data increases, with noticeable gains even on seen objects. These results underscore the importance of large-scale, diverse datasets for enhancing both performance and generalization in grasp learning, suggesting that dataset composition can be as influential as model architecture in determining overall effectiveness.

\noindent \textbf{Effect of the Number of Epochs.}
We conduct an ablation on the number of training epochs, and the results are presented in Figure~\ref{fig:data_and_epoch} (b). Performance on both seen and unseen objects increases steadily during the first five epochs. Beyond the sixth epoch, accuracy on seen objects saturates, while accuracy on unseen objects gradually declines, suggesting the onset of overfitting. This pattern illustrates a fundamental trade-off between optimizing in-distribution performance and maintaining out-of-distribution generalization in model learning of grasp generation.

\subsection{Generalization Study}
\label{subsec:generalization}

\begin{figure}[t]
\centering
\includegraphics[width=\columnwidth]{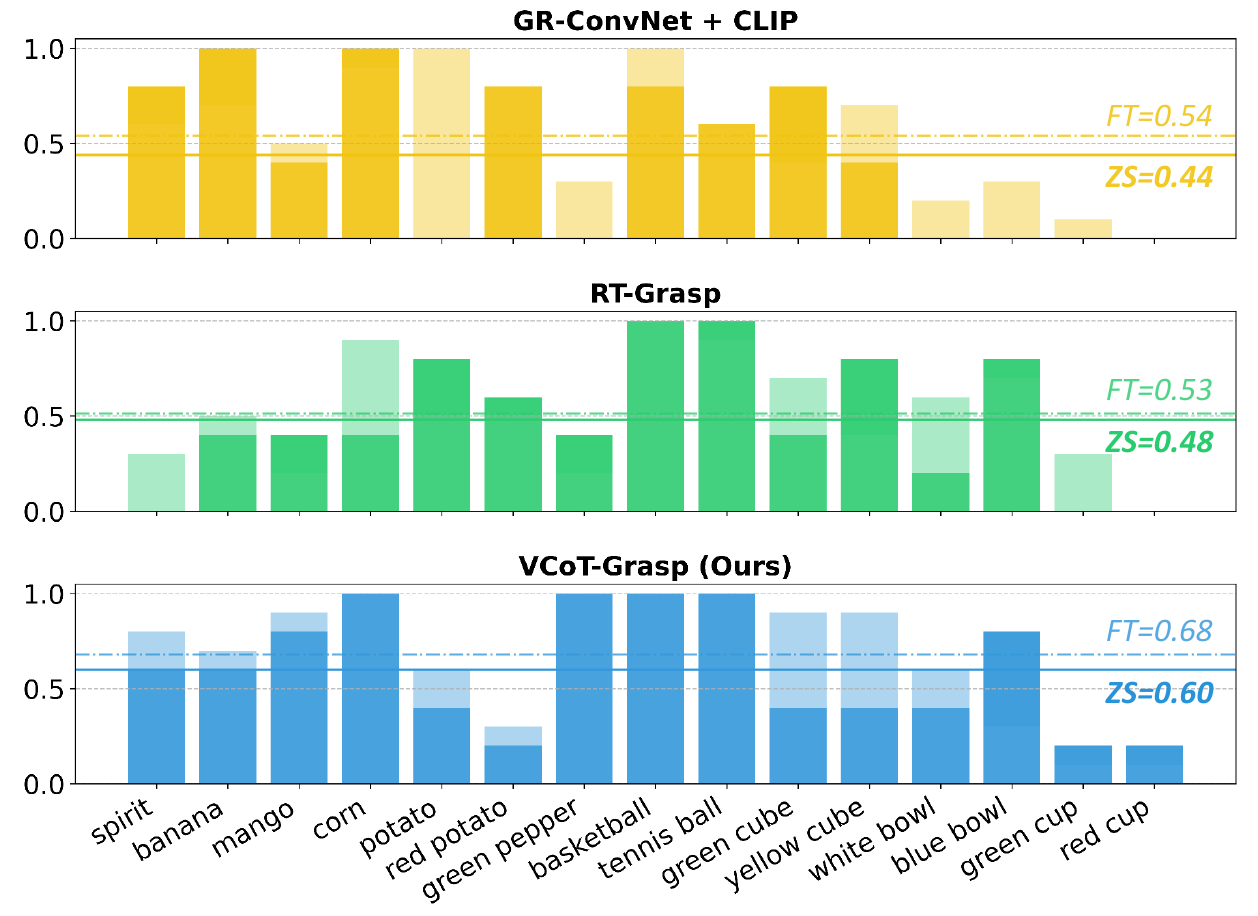}
\caption{Zero-shot performance on real-world unseen objects. Dark bars indicate zero-shot performance, while light bars represent the performance gains after fine-tuning on the training set.}
\label{fig:zero-shot}
\end{figure}

\noindent \textbf{Zero-shot Transfer.}
We also evaluate the zero-shot transfer ability of our method by directly deploying the model trained on VCoT-GraspSet to a real robot without any fine-tuning. As illustrated in Figure~\ref{fig:zero-shot}, the results on unseen objects demonstrate that VCoT-Grasp possesses strong transferability and generalization to real-world environments. In particular, it achieves a 12\% higher success rate than the state-of-the-art RT-Grasp, highlighting the effectiveness of incorporating visual chain-of-thought reasoning for bridging the gap between simulation and the real world.

\noindent \textbf{Robustness to Background Variations and Distractors.}
We further assess the robustness of our model by evaluating its generalization to complex environments with background variations and distractor objects, with representative examples shown in Figure~\ref{fig:exp_setup} (c). Five objects are selected, and each is tested over 10 trials. As reported in Table~\ref{table:generalization}, distractors degrade performance more than background changes. Even in these challenging settings, VCoT-Grasp maintains strong success rates. We attribute this robustness to its visual chain-of-thought reasoning, which encourages object-centric verification of graspable regions rather than reliance on global scene appearance, making it substantially less sensitive to clutter and background shifts.



\begin{table}[t]
\caption{Model robustness evaluation under background changes and additional distractors.}
\label{table:generalization}
\centering
\vskip -0.05in
\resizebox{0.5\textwidth}{!}{
\begin{tabular}{cccccc}
\toprule
Method & Background & Distractors & \makecell{Background \& \\ Distractors} \\
\midrule
GR-ConvNet~\cite{kumra2020antipodal} + CLIP~\cite{radford2021learning} & 29/50 & 25/50 & 20/50 \\
RT-Grasp~\cite{xu2024rt} & 27/50 & 25/50 & 21/50 \\
\rowcolor{gray!9.0} VCoT-Grasp (Ours) & \textbf{38/50} & \textbf{31/50} & \textbf{28/50} \\
\bottomrule
\end{tabular}
}
\vskip -0.05in
\end{table}

\section{Conclusion}

We propose VCoT-Grasp, a foundation model tailored for grasp generation. Unlike prior grasp foundation models that overemphasize dialog and textual reasoning while neglecting visual grounding, our approach introduces visual chain-of-thought reasoning to strengthen visual understanding. In addition, it helps fill an important gap by validating a multi-turn processing paradigm in robotic models. To support training, we refine and release a large-scale dataset, VCoT-GraspSet. Extensive experiments on both VCoT-GraspSet and real-world robot demonstrate the effectiveness of VCoT-Grasp, achieving superior in-distribution performance and significantly outperforming state-of-the-art baselines under diverse out-of-distribution scenarios, including unseen objects, domains, distractors, and background variations.




\section*{ACKNOWLEDGMENT}

This work was supported by the National Key R\&D Program of China (2023YFB4704900),
the Key Program of the Joint Funds of the National Natural Science Foundation of China (U25A20540),
and the Key Program of the National Natural Science Foundation of China (62436005).

\bibliography{main, IEEEabrv}

\end{document}